\pdfoutput=1

\documentclass[11pt]{article}

\usepackage[final]{acl}

\usepackage{times}
\usepackage{latexsym}
\usepackage{float}
\usepackage{subcaption}

\usepackage[T1]{fontenc}

\usepackage[utf8]{inputenc}

\usepackage{microtype}

\usepackage{inconsolata}

\usepackage{graphicx}

\title{miniGPT-Med: Large Language Model as a General Interface for Radiology Diagnosis}

\author{
 \textbf{Asma Alkhaldi\textsuperscript{1,2}\thanks{\texttt{\{asma.alkhaldi,mohamed.elhoseiny\}@kaust.edu.sa}}},
 \textbf{Raneem Alnajim\textsuperscript{1,2}},
 \textbf{Layan Alabdullatef\textsuperscript{1,2}},
 \textbf{Rawan Alyahya\textsuperscript{1}},
\\
 \textbf{Jun Chen\textsuperscript{2}},
 \textbf{Deyao Zhu\textsuperscript{2}},
 \textbf{Ahmed Alsinan\textsuperscript{1}},
 \textbf{Mohamed Elhoseiny \textsuperscript{2}\footnotemark[1]},
\\
\\
 \textsuperscript{1}Saudi Data and Artificial Intelligence Authority (SDAIA),\\
 \textsuperscript{2}King Abdullah University of Science and Technology (KAUST)
\\
}

\begin{document}
\maketitle
\begin{figure*}[t]
    \centering
\includegraphics[width=\textwidth]{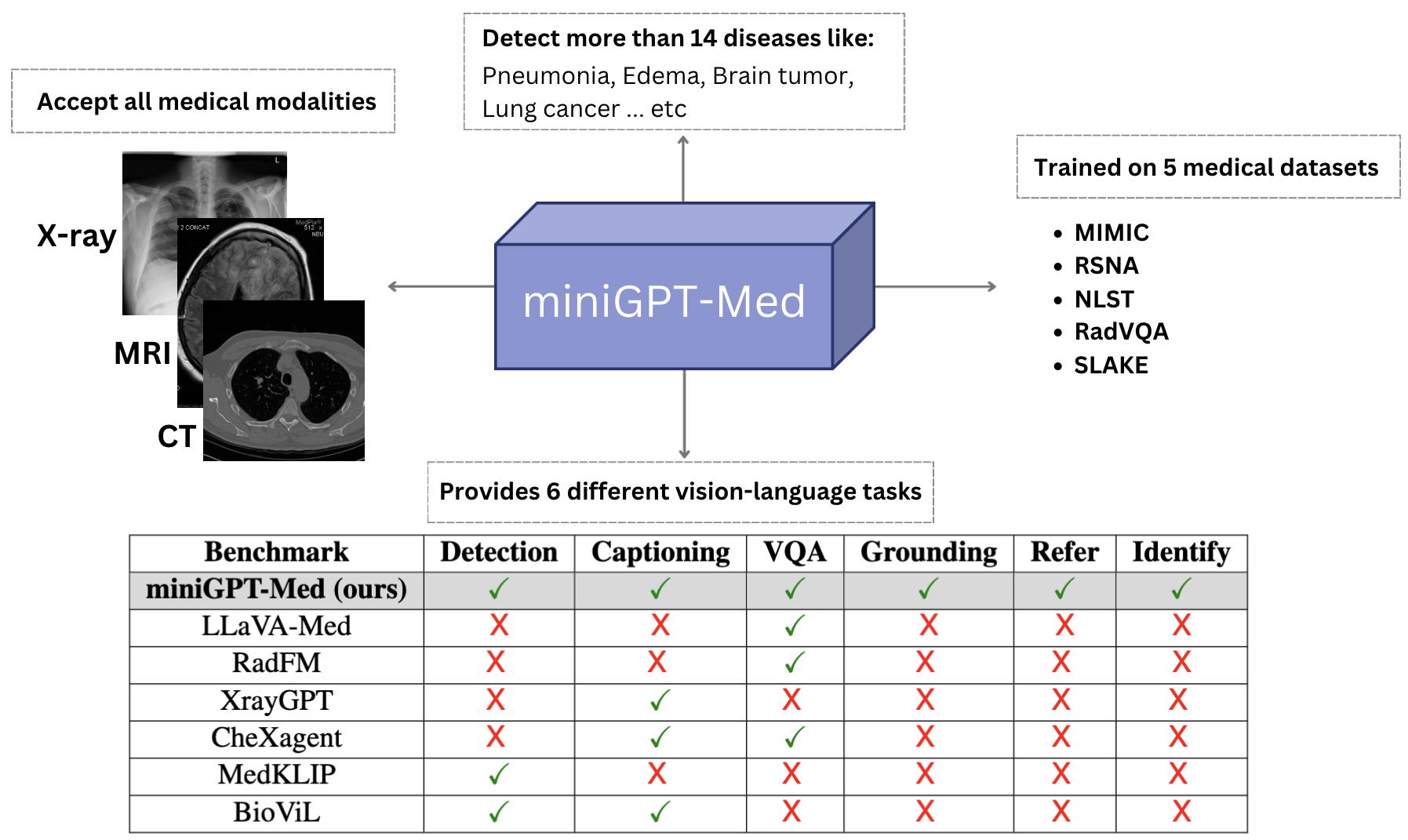}
\caption{The diverse capabilities by MiniGPT-Med. It can perform disease detection, medical visual question answering, and medical report generation. MiniGPT-Med effectively works with a wide range of radiological data (X-rays, CT scans, and MRIs) and is adept at diagnosing many diseases.}
    \label{fig:teaser}
\end{figure*}

\begin{abstract}
Recent advancements in artificial intelligence (AI) have precipitated significant breakthroughs in healthcare, particularly in refining diagnostic procedures. However, previous studies have often been constrained to limited functionalities. This study introduces MiniGPT-Med, a vision-language model derived from large-scale language models and tailored for medical applications. MiniGPT-Med demonstrates remarkable versatility across various imaging modalities, including X-rays, CT scans, and MRIs, enhancing its utility. The model is capable of performing tasks such as medical report generation, visual question answering (VQA), and disease identification within medical imagery. Its integrated processing of both image and textual clinical data markedly improves diagnostic accuracy. Our empirical assessments confirm MiniGPT-Med's superior performance in disease grounding, medical report generation, and VQA benchmarks, representing a significant step towards reducing the gap in assisting radiology practice. Furthermore, it achieves state-of-the-art performance on medical report generation, higher than the previous best model by 19\% accuracy. MiniGPT-Med promises to become a general interface for radiology diagnoses, enhancing diagnostic efficiency across a wide range of medical imaging applications. Our model and code have been made publicly available \href{https://github.com/Vision-CAIR/MiniGPT-Med}{https://github.com/Vision-CAIR/MiniGPT-Med}
\end{abstract}

\section{Introduction}
The unprecedented surge in both the quantity of image-text data across diverse fields and the strides made in vision-language modeling have paved the way for groundbreaking research in Generative Pretraining. This era of innovation is marked by the emergence of multimodal models such as GPT-4 \cite{achiam2023gpt} and Gemini \cite{team2023gemini}. These advancements signify a leap forward in our ability to process and understand complex data. Despite this progress, the adoption of Multi-modal Large Language Models (LLMs) within the medical sector remains limited. 
The medical field's unique requirements for data complexity, sensitivity, and specificity highlight the need for tailored approaches to harness the potential of LLMs in transforming healthcare research and practice. Numerous models designed for medical applications have been introduced, yet they often exhibit a high degree of specialization for specific tasks. This specialization limits their versatility, particularly in performing diverse medical applications. For instance, models like Med-Flamingo \cite{medflamingo} and XrayGPT \cite{thawkar2023xraygpt} are primarily tailored for tasks such as medical report generation and medical visual question answering, respectively. However, they lack capabilities in essential areas like disease detection, which requires visual grounding skills— a crucial component in the medical field. To address this deficiency, we introduce MiniGPT-Med, a unified model capable of adeptly handling both grounding and non-grounding tasks. We introduce MiniGPT-Med, a versatile model designed for various tasks in the medical domain, including but not limited to medical report generation, medical visual question answering, and disease identification. MiniGPT-Med builds upon the architecture of large language models (LLMs), which have demonstrated exceptional generative capabilities and extensive linguistics, including medical knowledge. Drawing on the successes of LLMs in a wide range of vision-language applications, as evidenced in recent studies \cite{zhu2023minigpt,chen2023minigpt,li2024llava}, our model adopts a design similar to MiniGPT-v2 \cite{chen2023minigpt}, utilizing the LLaMA-2 language model as a universal interface. Additionally, we incorporate distinct task identifiers to enhance the model's ability to accurately perform various medical vision-language skills. Through extensive experimentation, we have demonstrated that our model exhibits strong performance across a range of medical vision-language tasks, including medical report generation, medical visual question answering, and disease detection. We benchmarked our model against both specialized and generalized baseline models, revealing that our approach achieves strong results across all evaluated tasks. Notably, in the domain of medical report generation, our model attained state-of-the-art performance, surpassing the best baseline models by 19\% in BERT-Sim and 5.2\% in CheXbert-Sim. This indicates our model has strong generation capabilities on diverse medical vision-language tasks.

\noindent Our contributions are as follows:

\begin{enumerate}
    \item We introduce MiniGPT-Med, a model tailored for the heterogeneous nature of radiological imagery, encompassing X-rays, CT scans, and MRIs. This model is adept at handling a variety of vision-language tasks, including disease identification, medical visual question answering, and the generation of medical reports.
    \item Through comprehensive evaluation, we evaluated our model across both grounding and non-grounding tasks, complemented by expert manual assessments. The findings demonstrate that MiniGPT-Med delivers competitive performance across a majority of benchmarks, surpassing both generalized and specialized models, notably achieving state-of-the-art results in medical report generation, and surpassing the best baseline by 19.0\%.
\end{enumerate}

\section{Background}
\textbf{Aligning Visual Data with Large Language Models:} Recent advancements in the domain of large language models such as the release of GPT-4, have enhanced the interpretative and generative capabilities of LLMs. This progress is exemplified by models such as LLaVA\cite{liu2023visual}, Flamingo\cite{alayrac2022flamingo}, and MiniGPT-v2\cite{chen2023minigpt}. LLaVA is designed to augment the understanding of visual content in large language models through diverse multimodal instructions. This enhancement in comprehension is critical for integrating different forms of data input. In contrast, Flamingo demonstrates remarkable proficiency in quick adaption to novel tasks with minimal data. This model effectively manages sequences that incorporate both visual and textual elements. MiniGPT-v2, on the other hand, displays enhanced multimodal capabilities within a singular model framework. This is achieved through task-specific training and a specialized architecture that combines visual tokens with a large language model, aligning well with the objectives of LLaVA and Flamingo.\\

\textbf{Integration of Vision Language Models for Enhanced Medical Diagnostics:} Recent work in vision-language models has led to significant improvements in healthcare applications, especially in medical image analysis and diagnostic report generation. Utilizing VLMs in medical diagnostics marks a significant progression in the healthcare industry. models combine computer vision and language processing to better analyze medical images like X-rays, computed tomography (CT), and MRIs. More specialized applications in the medical field such as LLaVA-Med \cite{li2024llava} and Med-BERT \cite{MedBERT} have shown promise in incorporating structured electronic health records for improvements in disease prediction tasks. MedVQA \cite{MedVQA} has demonstrated medical visual question-answering and image analysis capabilities. Furthermore, for classification and interpretation tasks, Med-Flamingo \cite{medflamingo}, MedVis \cite{shen2008medvis}, and MedMCQA \cite{MedMCQA} have shown the importance of few-shot learning, visual interpretation, and domain-specific question-answering in medical AI. Both LLaVA-Med and Med-Flamingo focus on multimodal conversational AI and few-shot learning in medical contexts, utilizing large-scale datasets and showcasing proficiency in visual question answering. BioViL \cite{bannur2023learning}, BioBERT \cite{BioBERT}, and BioGPT \cite{BioGPT} all have tackled a more domain-specific language model pretraining. BioViL emphasizes text semantics for enhanced biomedical vision-language processing. Emphasis on specialized models for radiology applications has also been presented in MedKLIP \cite{wu2023medklip}, XrayGPT \cite{XrayGPT}, and BERTHop  \cite{monajatipoor2021berthop} all demonstrating the challenge of achieving high diagnostic accuracy. MedKLIP in particular innovates by integrating medical knowledge into vision-language pre-training for improved disease classification. XrayGPT integrated a medical visual encoder with a large language model to combine visual and textual analysis to generate precise summaries from radiological data, while BERTHop showed diagnostic performance with smaller datasets on chest X-rays. Moreover, the contributions of CheXagent \cite{chen2024chexagent}, CheXNeXt \cite{CheXNeXt}, and CheXpert \cite{irvin2019chexpert} have set benchmarks in chest pathologies detection. While each work presents unique approaches, their common goal is to enhance radiological analysis through sophisticated AI models.

\section{Method}
\subsection{Model architecture}
Our model architecture, illustrated in Fig \ref{fig:arch}, is composed of three key components: a visual backbone, a linear projection layer, and an extensive language model. The details of each component are described as follows:

\begin{figure*}[t]
    \centering
\includegraphics[width=\textwidth]{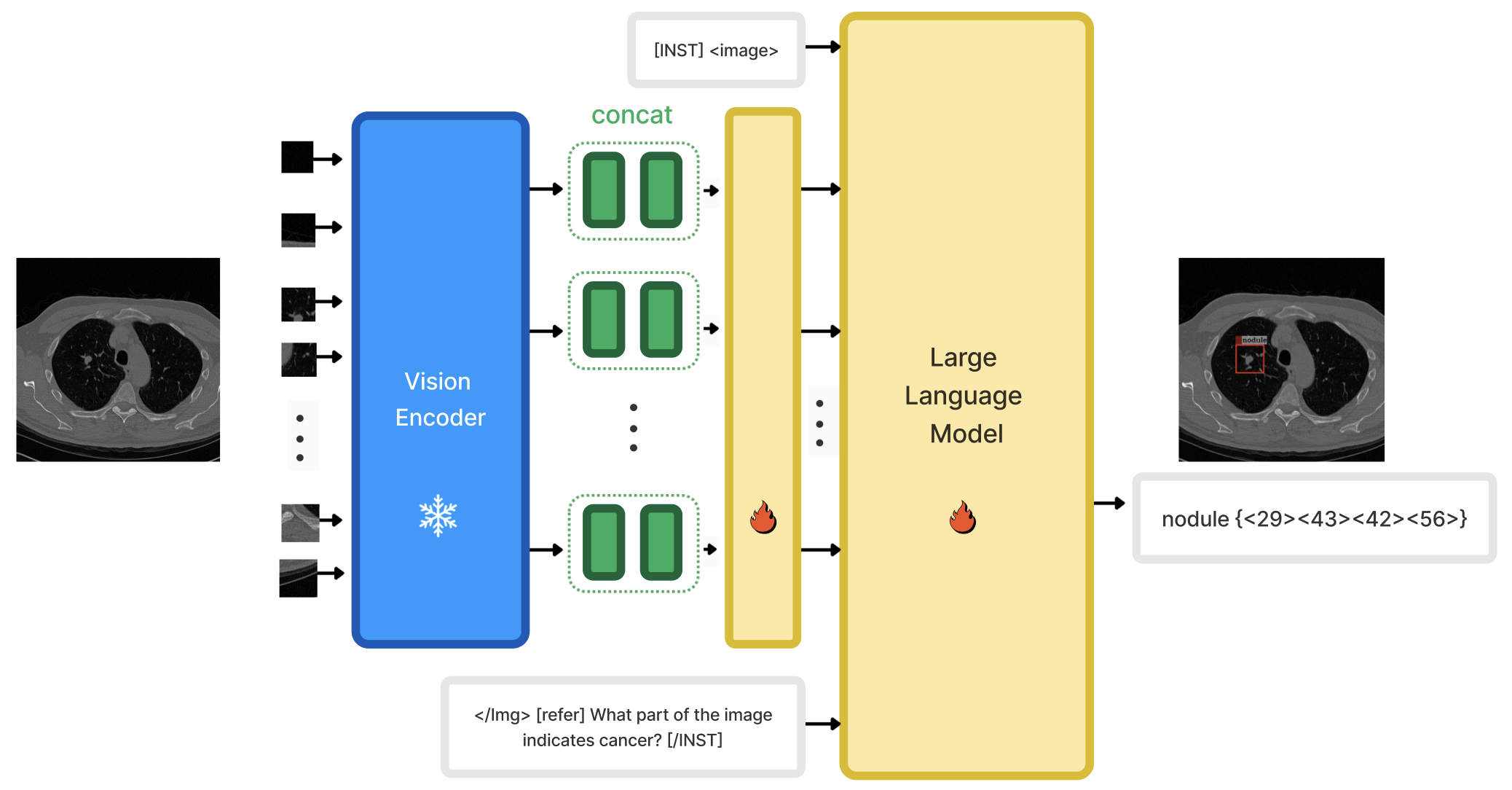}
\caption{MiniGPT-Med Architecture Overview: The architecture comprises a vision encoder, a linear projection layer, and a large language model. It processes a single medical image, transforming it into visual semantic features via a pre-trained vision encoder. These features are concatenated into a single visual token. A linear projection layer then maps these visual tokens into the large language model's space. Throughout the training process, we maintain the vision encoder's parameters constant while fine-tuning the large language model and linear projection layer.}
    \label{fig:arch}
\end{figure*}

\noindent \textbf{Vision Encoder.} In our approach, we incorporate EVA \cite{sun2023eva} as the primary visual backbone of our model. EVA \cite{sun2023eva}, a high-performing vision encoder, can be particularly effective when applied to radiological data due to its ability to handle complex image structures and variations. Throughout the entire training process, this visual backbone remains frozen during the training. The radiological images are usually in high resolution, we train the models with the image resolution of 448$\times$448. We also interpolate the positional encoding to adapt to the higher image resolution.

\noindent \textbf{Large Language Model (LLM).} We have incorporated the LLaMA2-chat (7B)\cite{touvron2023llama2}, an open-source language model, as the primary language model backbone. The LLM has already learned extensive medical knowledge through learning huge linguistic knowledge, and we treat it as a unified interface for processing many medical vision-language tasks. For example, the LLM can help generate detailed medical reports and also the precise localization of tumors in the medical domain.

\noindent \textbf{Vision Language Alignment.} We adopt the architecture of MiniGPT-v2 \cite{chen2023minigpt} and enhance efficiency by concatenating visual tokens from the vision encoder, a technique particularly beneficial for processing high-resolution medical images. This method involves merging four adjacent visual tokens into a single embedding, which is then mapped into the language model's feature space via a linear projection layer.

\subsection{Prompt Template.}
We employ a prompt template that allows our model can well deal with many diverse medical vision-language skills, such as visual question answering, image captioning, referring expression comprehension (REC), referring expression generation (REG), disease detection, and grounded image captioning. A language model might experience high levels of hallucination and confusion while dealing with many diverse vision-language tasks. For instance, when being asked to identify a potential lung tumor, it could mistakenly focus on and describe areas of calcification in the blood vessels or heart. Therefore, to avoid ambiguity in these multi-task environments, we add task-specific tokens to our training framework. We follow a similar instruction design to that of MiniGPT-v2\cite{chen2023minigpt} in our instruction template, presented as follows:
\begin{center}
\small
\textit{[INST] $<$Img$>$ $<$ ImageFeature$>$ $<$/Img$>$ [Task Identifier] Instruction [/INST]} 
\end{center}

We present diverse prompt templates in Table \ref{tab:instructions} to demonstrate how our model effectively deals with the different tasks through task identifiers.


\subsection{Region grounding representation.} 
For grounding skills that involve the spatial location of objects, such as disease detection and grounded image captioning, we employ a textual representation for bounding boxes. This representation allows us to integrate spatial locations into the text fed into the language model. We normalize the bounding box coordinates within the [0,100] range. Each spatial location is expressed in the format:
\[
\{
<X_{left}>
<Y_{top}>
<X_{right}>
<Y_{bottom}>
\}
\]

\section{Experiments}
The experiment aimed to evaluate the efficacy of MiniGPT-Med with a focus on its ability to accurately analyze and describe complex medical imaging data for applications like lung cancer detection, report generation, and question-and-answer capabilities. We fine-tuned stage 3 of MiniGPT-v2 using a comprehensive dataset of radiological images, including X-rays, MRIs, and CT scans, covering a wide range of medical conditions for a variety of skills.

\subsection{Dataset Setup}
The lack of quality medical datasets is a significant challenge in the field of deep learning for medical imaging. To address this issue, we have prioritized the collection of a comprehensive dataset focusing on radiology, specifically lung diseases, as well as general medical information. Our goal is to gather a diverse and extensive range of medical images, including X-rays, CT scans, and MRI images. Furthermore, we aim to enhance the dataset by incorporating images with bounding boxes, datasets featuring question-and-answer formats, and datasets for report generation. These additions will support all the necessary skills for the model training and development.

\begin{table*}[t]
  \centering
  \begin{tabular}{lc}
    \hline
    \textbf{Task Types} & \textbf{[Identifier] Instruction} \\
    \hline
    Caption & [caption] Could you describe the contents of this image for me?\\
    VQA & [vqa] What plane is the image in?\\
    Detection & [detection] pneumonia \\
    Refer & [refer] the nodule in the left lung\\
    Grounding & [grounding] describe this image in detail \\
    Identify &  [identify] what is this \{ \textless{}56\textgreater{}\textless{}16\textgreater{}\textless{}84\textgreater{}\textless{}58\textgreater{}\} \\
    \hline\\
  \end{tabular}
  \caption{Task-specific instruction format. \textless{}ImageFeature\textgreater{} denote the image features. During our model training, we used six different types of task identifiers for diverse grounding and non-grounding tasks.}
  \label{tab:instructions}
\end{table*}

The collected datasets include
MIMIC \cite{johnson2019mimic}, NLST \cite{tciaNLST} and SLAKE \cite{slake}, RSNA \cite{rsna2018pneumonia} and RadVQA \cite{osfProject}.The details for those medical datasets are demonstrated as the following: 
    \noindent \textbf{MIMIC} The dataset comprises 377,110 images and 227,835 medical reports. In our study, we obtained the preprocessed MIMIC dataset from XrayGPT \cite{thawkar2023xraygpt}, which includes 114,539 de-identified chest X-ray images in JPG format, each accompanied by a corresponding radiology report. Of these, 171,085 images and reports are allocated for training, while 43,454 images and reports are designated for testing. This dataset is utilized for the task of report generation.

    \noindent \textbf{NLST} This dataset is employed for the detection task, encompassing 7,625 meticulously annotated low-dose CT scan images for lung cancer, specifically marked to pinpoint nodule locations. From the complete 3D volume, we extracted the 2D CT slice displaying the nodule. These annotations, utilized for training, were sourced from the work of Sybil \cite{mikhael2023sybil}. 

    \noindent \textbf{SLAKE} This dataset is used for training the grounding and VQA tasks, where it comprises 579 radiology images delineating various body organs, coupled with 3,543 diverse sets of question-answer pairs used for training. 
    
    \noindent \textbf{RSNA.} We use the RSNA dataset for evaluating pneumonia detection task. RSNA dataset comprises 1,218 patients who had at least one or more pneumonia conditions. We perform the zero-shot evaluation on this dataset for the disease detection task.
    
    \noindent \textbf{RadVQA} Includes 315 radiology images evenly spread across the head, chest, and abdomen, each paired with multiple questions and results in 2,248 question-answer pairs. These questions fall into 11 distinct categories: abnormality, attribute, modality, organ system, color, counting, presence of objects or conditions, size, plane, and positional reasoning. Half of the responses are closed-ended (i.e., yes/no), while the remaining are open-ended, typically requiring one-word or short-phrase replies. We perform the zero-shot evaluation on RadVQA datasets.

\subsection{Training Details}
In our experiment, we initialize our model with MiniGPT-v2 \cite{chen2023minigpt} pre-trained weights (after stage 3) and keep the vision encoder frozen throughout the whole training process. We finetune the linear projection layer and use LoRA (Low-Rank Adaptation) \cite{lora} to finetune the LLaMA-2 \cite{touvron2023llama2} large language model. The model is trained using the cross-entropy loss function, which is optimized using the AdamW optimizer. Our dataset comprises 124,276 medical images, each with a resolution of 448x448 pixels, and no data augmentation is applied. The entire training was performed on a single NVIDIA A100 GPU over 100 epochs, with a maximum learning rate of 1e-5. The training duration was approximately 22 hours.

\subsection{Baseline models}
In this study, we conducted an assessment of MiniGPT-Med's performance across three distinct tasks: medical report generation, disease detection, and medical visual question answering (VQA). We compared our model to both specialist and generalist models. The specialist models represent those who can only do either grounding or non-grounding tasks. The generalist models represent those models that can do various tasks including both grounding and non-grounding tasks.

\noindent -- For the \textbf{medical report generation} task, we compared MiniGPT-Med with specialist models including Med-Flamingo \cite{medflamingo} and LLaVA-Med \cite{li2024llava}, known for their prowess in vision-language tasks and contextual learning abilities. Additionally, we compared MiniGPT-Med with RadFM \cite{wu2023towards}, which is specifically tailored for radiology, and XrayGPT \cite{thawkar2023xraygpt}, a novel vision-language model designed for chest radiograph analysis. Furthermore, we evaluated MiniGPT-Med against CheXagent \cite{chen2024chexagent}, a foundation model focused on improving chest X-ray interpretation. Moreover, comparisons were made with generalist models like MiniGPT-v2 and Qwen-VL \cite{bai2023qwen}, trained on the general vision-language data, showcasing exceptional performance across various vision-focused comprehension benchmarks. 

\noindent -- In the  \textbf{disease detection} task, MiniGPT-Med was compared against specialist models including BioVil \cite{bannur2023learning}, MedKLIP \cite{wu2023medklip}, and GLoRIA \cite{huang2021gloria}, all pre-trained on vision-language medical datasets, as well as generalist models including MiniGPT-v2 and Qwen-VL. 

\noindent -- In the \textbf{medical VQA} task, we compared MiniGPT-Med with specialized models like MedVINT \cite{zhang2023pmc}, OpenFlamingo \cite{awadalla2023openflamingo}, and Med-Flamingo \cite{medflamingo} tailored to address the challenges of medical VQA, particularly in zero-shot scenarios, utilizing the RadVQA dataset. Additionally, our work was compared with generalist models such as MiniGPT-v2 and Qwen-VL to provide a comprehensive evaluation of MiniGPT-Med's performance.

\begin{table*}[t]
  \centering
  \begin{tabular}{lcccc}
    \hline
    \textbf{Method} & \textbf{Model's type} & \multicolumn{2}{c}{\textbf{MIMIC-CXR}} \\
    \cline{3-4}
    & & BERT-Sim & CheXbert-Sim \\
    \hline
    MedFlamingo &  & 10.4 & 3.2 \\
    LLaVA-Med &  & 6.2 & 17.5 \\
    RadFM & Specialist Models & 45.7 & 17.5 \\
    XrayGPT &   & 44.0 & 24.2  \\
    CheXagent  &   & \textbf{50.4} & \textbf{24.9} \\
    \hline
    MiniGPT-v2 &   & 53.0 & 21.1 \\
    Qwen-VL & Generalist Models & 51.9 & 20.3 \\
    \textbf{Ours} &   & \textbf{72.0} & \textbf{30.1} \\
    \hline
  \end{tabular}
  \caption{Evaluation of Medical Report Generation: MiniGPT-Med versus Generalist and Specialist Models. MiniGPT-Med is contrasted with a generalist model capable of executing a wide range of grounding and non-grounding tasks, alongside specialist models limited to non-grounding tasks. The highest performance metrics for both specialist and general models are highlighted in \textbf{bold}.}
  \label{tab:mimic_comparison}
\end{table*}

\begin{table}[t]
  \centering
  \begin{tabular}{p{0.35\columnwidth} p{0.33\columnwidth} p{0.18\columnwidth}}
    \hline
    \textbf{Method} & \textbf{Model's type} & \textbf{RSNA IoU} \\
    \hline
    BioViL  &   & 0.30  \\
    MedKLIP   & Specialist & \textbf{0.31} \\
    GLoRIA  &   & 0.21 \\
    \hline 
    Qwen-VL &   & 0.10\\
    MiniGPT-v2 & Generalist & 0.13 \\
    \textbf{Ours} &   & \textbf{0.26} \\
    \hline
  \end{tabular}
  \caption{Evaluation of Disease Detection on the RSNA Benchmark as zero-shot: A Comparison of Our Models with Generalist and Non-Generalist Models. The top performance metrics for both specialist and general models are highlighted in \textbf{bold}.}
  \label{rsna_comparison}
\end{table}

\subsection{Evaluation Metrics}
In our study, we adapted our evaluation approach to align with the distinct skills required for interpreting radiology images using MiniGPT-Med. To assess the model's ability to generate radiological reports, we used two metrics: BERT Similarity (BERTsim) and CheXbert Similarity (CheXbert-Sim). BERTsim was utilized to evaluate the semantic similarity between the model-generated descriptions of radiological images and the expert-provided ground truth annotations. This involved using a BERT model to embed both the ground truth and generated sentences, followed by computing the cosine similarity between these embeddings. CheXbert-Sim, conversely, was selected for its relevance in assessing the model's accuracy in replicating professional medical report standards. It is a specialized version of the BERT model, fine-tuned on clinical texts, which computes the cosine similarity between embeddings for each corresponding sentence pair after encoding. For the Visual Question Answering (VQA) aspect, we exclusively used BERTsim to measure the semantic accuracy of the model’s responses. Additionally, we employed Intersection over Union (IoU) for grounding, a metric that quantitatively measures the model’s precision in localizing and identifying specific features or abnormalities within the radiology images, such as pneumonia in the RSNA dataset.

\subsection{Medical Report Generation} 
In our comprehensive study, we evaluated the efficacy of the MiniGPT-Med model in the generation of medical report generation, leveraging the comprehensive MIMIC dataset \cite{johnson2019mimic}. The results of this evaluation, which are outlined in Table \ref{tab:radiologist_comparison}, demonstrate that the MiniGPT-Med model surpasses both specialized and generalized baseline models. Most notably, MiniGPT-Med demonstrates a significant edge over the leading specialized model, CheXagent \cite{chen2024chexagent}, with remarkable margins of 21.6 and 5.2 on the BERT-Sim and CheXbert-Sim metrics, respectively. This performance not only showcases MiniGPT-Med's supremacy in the medical report generation but also underscores its ability to outpace the top generalist models substantially—by a notable 19 points on BERT-Sim and 9 points on CheXbert-Sim. These findings solidify MiniGPT-Med's position as a cutting-edge tool, demonstrating its effectiveness in medical report generation.

\subsection{Disease Detection} 
The data showcased in Table \ref{rsna_comparison} reveal that MiniGPT-Med stands out for its competitive performance when compared against a comprehensive range of baseline models. With an Intersection over Union (IoU) score of 0.26, MiniGPT-Med not only exceeds the capabilities of generalist models by a margin of 16\% but also attains performance metrics on par with specialist models. The peak IoU score among these specialist models is noted to be 0.31. Our MiniGPT-Med achieves competitive results and it demonstrates good disease detection performance among all the baseline models, highlighting its potential as a versatile and effective tool in the medical domain.

\subsection{Medical Visual Question Answering}
This study evaluates our model, MiniGPT-Med, against various baseline models using the RadVQA \cite{osfProject} benchmark, as presented in Table \ref{radvqa_comparison}. MiniGPT-Med achieves a notable performance metric of 0.58, surpassing both generalist models such as MiniGPT-v2 \cite{chen2023minigpt} and specialist models like OpenFlamingo \cite{awadalla2023openflamingo} and Med-Flamingo \cite{medflamingo}. This performance not only demonstrates MiniGPT-Med's superiority over a broad range of models but also shows it can achieve results comparable to those of the leading specialist model, MedVIN \cite{zhang2023pmc}, which has an accuracy of 0.62. The ability of MiniGPT-Med to outperform or match the performance of several specialized and generalist models underscores its significant potential as a foundation for the development of advanced medical visual question-answering models.

\begin{table}[t]
  \centering
  \begin{tabular}{p{0.34\columnwidth} p{0.30\columnwidth} p{0.19\columnwidth}}
    \hline
    \textbf{Method} & \textbf{Model's type} & \textbf{RadVQA} \\
     &   & \textbf{\small BERT-Sim}\\
     \hline
    MedVIN &   & \textbf{0.62} \\
    OpenFlamingo & Specialist & 0.49 \\
    Med-Flamingo&   & 0.48 \\
    \hline
    Qwen-VL &  & 0.13 \\
    MiniGPT-v2 & Generalist & 0.55 \\
    \textbf{Ours}  &  & \textbf{0.58} \\
    \hline
  \end{tabular}
  \caption{Evaluation of visual Question Answering on the radVQA Benchmark as zero-shot: A Comparison of Our Models with Generalist and Non-Generalist Models. The top performance metrics for both specialist and general models are highlighted in \textbf{bold}.}
  \label{radvqa_comparison}
\end{table}

\subsection{Radiology Expert Evaluation}
Our study evaluated MiniGPT-Med using a rigorous human subjective protocol with two senior radiologists. They assessed 50 random samples from the MIMIC dataset's test suite, focusing on the model's robustness, granularity, and accuracy. The evaluation centered on three questions, Q1: how closely does the generated report align with your expert judgment? Q2: How detailed is the medical content of the generated report? Q3: How accurate is the generated report in diagnosing pathologies?
We present the results in the accompanying Table \ref{tab:radiologist_comparison}. 
The results show that a remarkable 76\% of the artificial medical reports are adjudged as of high quality. A further 19\% were classified as of medium quality, while a mere 5\% were deemed to be of poor quality. This distribution underscores the model's capability to synthesize medical reports that not only resonate with professional standards but also exhibit a high degree of detail and diagnostic accuracy. Such findings underscore the potential of MiniGPT-Med as a valuable tool in the augmentation of medical reporting processes, indicating its substantial reliability and effectiveness in generating clinically relevant reports.
\begin{table}[H]
    \centering
    \begin{tabular}{@{}lcc@{}}
        \hline
        \textbf{ } & \multicolumn{1}{c}{\textbf{Radiologist Evaluation}} & \\
        \textbf{Quality} & Percentage & \\
        \hline
        Good   & 76\%  &\\
        Medium & 19\%    & \\
        Poor   &  5\% & \\
        \hline
    \end{tabular}
    \caption{Expert Manual Evaluation for Medical Report Generation. We evaluate the model in terms of robustness, granularity, and accuracy. The table presents the percentage of votes for different quality categories.}
    \label{tab:radiologist_comparison}
\end{table}

\subsection{Qualitative Evaluation}
In this section, we provide comprehensive demonstrations of MiniGPT-Med's capabilities in generating medical reports and performing interpretative tasks. First, Figure \ref{fig:report_generation} illustrates the model's ability to produce detailed medical reports from imagery data. Also, the model can accurately identify and delineate specified abnormalities with bounding boxes as shown in Figure \ref{fig:detection}. Additionally, Figure \ref{fig:grounding} demonstrates the grounding skill, where the model explains each generated word and draws a bounding box around the object. Furthermore, Figure \ref{fig:refer} details the model's precision in referencing, and pinpointing abnormalities as specified by users. Moreover, the identification feature is showcased in Figure \ref{fig:identify}, where the model provides elaborate medical descriptions utilizing object coordinates. Finally, Figure \ref{fig:vqa} presents the model's visual question-answering (VQA) functionality, underscoring its effectiveness in providing precise answers to medical questions.


\section{Limitation}
MiniGPT-Med faces challenges due to a lack of diverse and high-quality training datasets, limiting its coverage to a narrow range of diseases. To improve, richer and more diverse datasets are needed, along with advanced vision backbones and enhancements in the underlying large language model. The model occasionally generates inaccurate medical reports and improperly connects symptoms to diseases, a phenomenon known as hallucination. Additionally, it struggles to distinguish between the abnormality and the medical images that include device implants in the human body. Fig.\ref{fig:limitation} demonstrates a sample of the data that MiniGPT-Med failed to correctly identify the pneumonia location. The object under the green bounding box is the ground truth and the object under the red bounding box is the false detection. The model easily confuses the device implants as an abnormality. This shortcoming often results in misdiagnosed conditions. Specifically, when AI encounters X-rays or MRIs featuring implants, it may incorrectly identify these as abnormalities.

\section{Conclusions}
In this study, we introduce MiniGPT-Med, a specialized multi-modal designed for radiology diagnosis applications. It handles various medical vision-language tasks such as generating medical reports, detecting diseases, and answering visually-based medical questions, using distinct task identifiers to navigate these tasks efficiently. MiniGPT-Med outperforms baseline models in both grounding and non-grounding tasks, achieving state-of-the-art performance in the MIMIC-CXR medical report generation task. Radiologist evaluations show that approximately 76\% of the generated reports are of preferred quality, highlighting the model's superiority. Future plans include incorporating more diverse medical datasets, improving the understanding of complex medical terminology, enhancing interpretability and dependability, and conducting extensive clinical validation studies to ensure effectiveness and safety in real healthcare environments.

\section*{Acknowledgments}
This work was supported by KAUST-SDAIA funding. Asma, Raneem, and Layan started working on this project as visiting research engineers at KAUST VisionCAIR. We extend our sincere gratitude to the KAUST HPC/Ibex cluster Team for their invaluable assistance and support during this research project to train the model.


\bibliography{custom}
\appendix
\section{Qualitative Examples} 

\begin{figure}[H]
    \centering
    \begin{subfigure}[b]{0.40\textwidth}
        \centering
        \includegraphics[width=\textwidth]{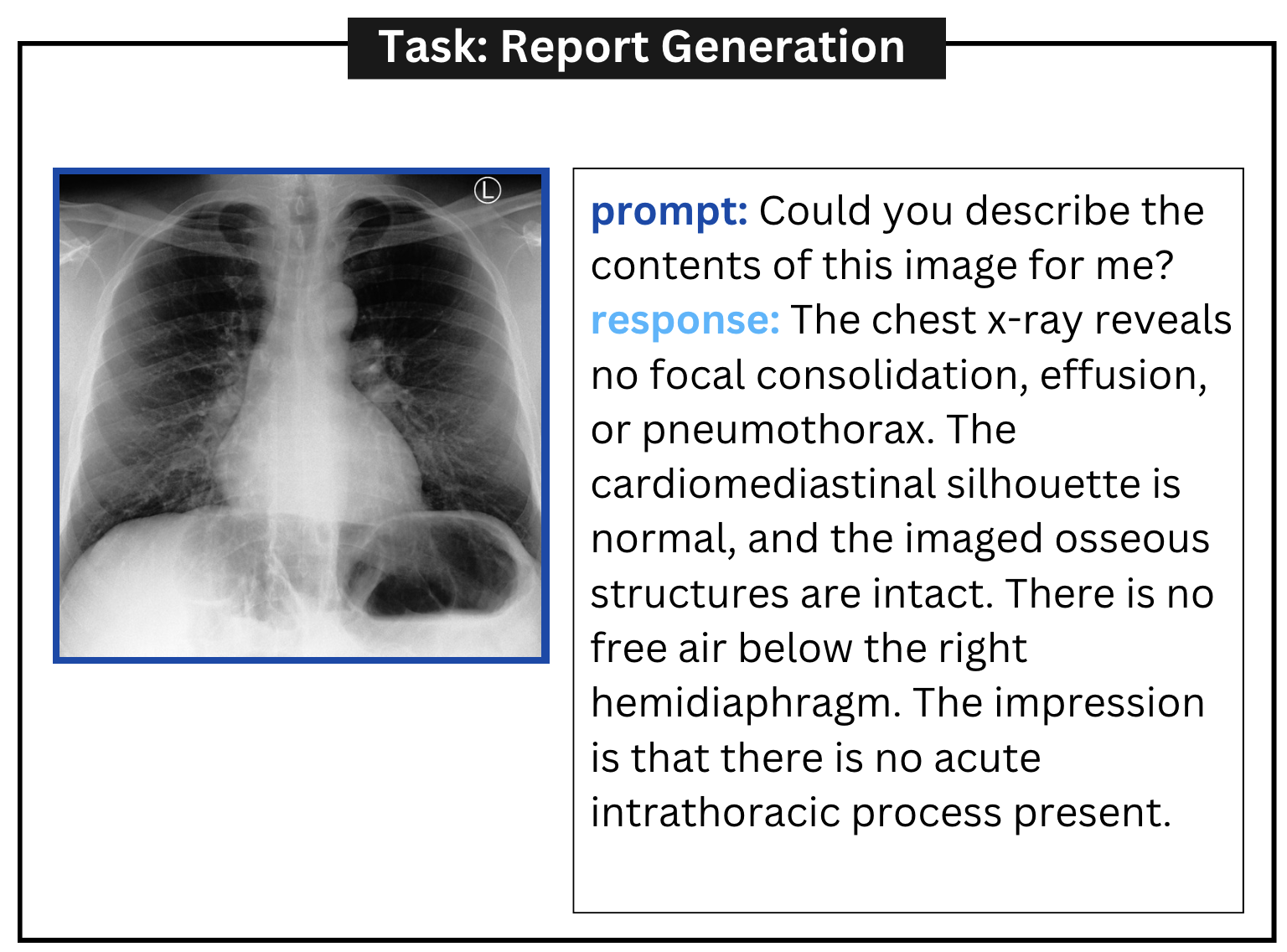}
        \caption{Report Generation Task}
        \label{fig:report_generation}
    \end{subfigure}
    \hfill
    \begin{subfigure}[b]{0.40\textwidth}
        \centering
        \includegraphics[width=\textwidth]{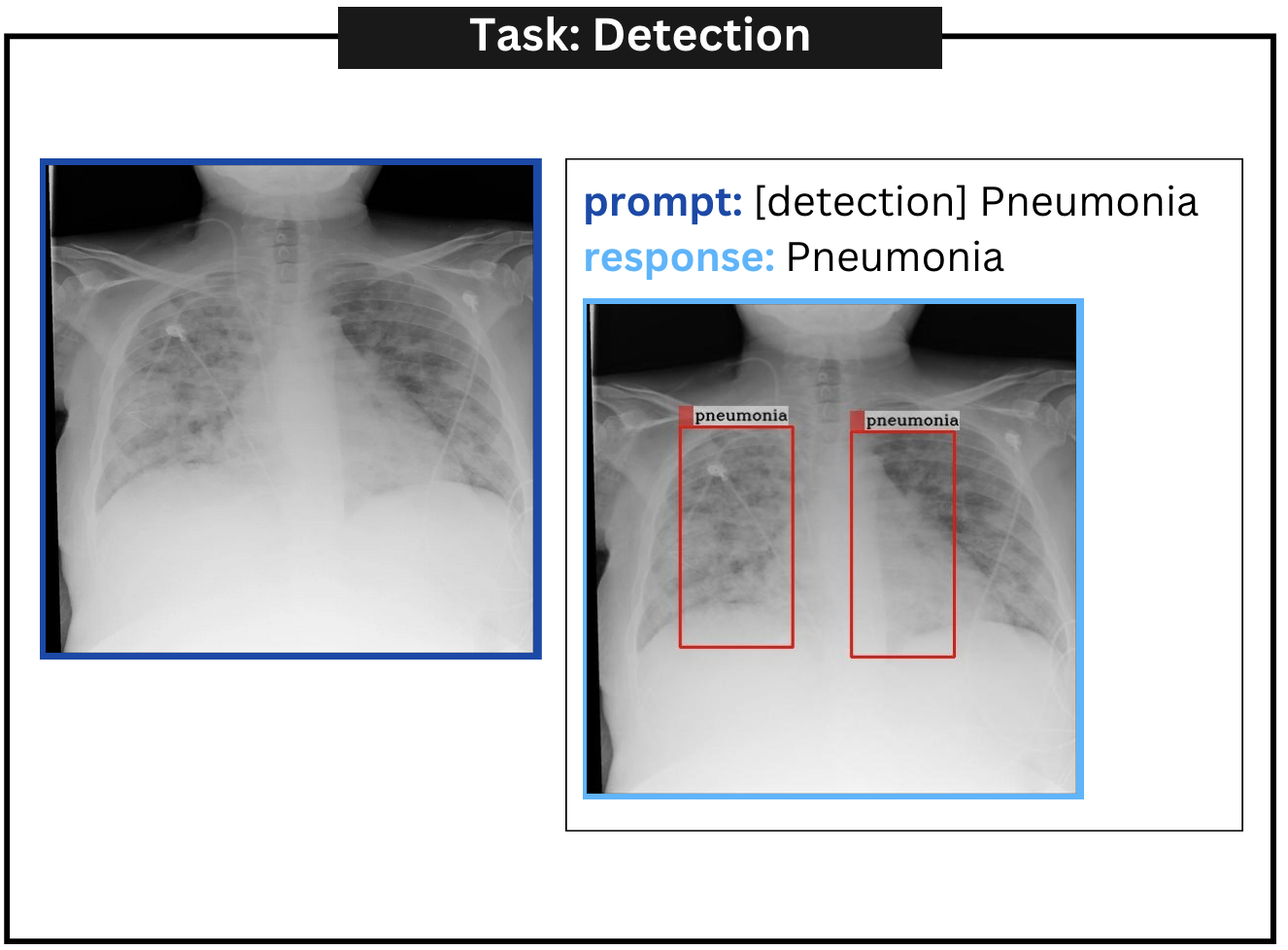}
        \caption{Detection Task}
        \label{fig:detection}
    \end{subfigure}
\caption{Examples of MiniGPT-Med multi-task abilities include: (a) medical report generation and (b) disease detection}
\label{fig:tasks}
\end{figure}

\begin{figure}[H]\ContinuedFloat
    \centering 
    \begin{subfigure}[b]{0.43\textwidth}
        \centering
        \includegraphics[width=\textwidth]{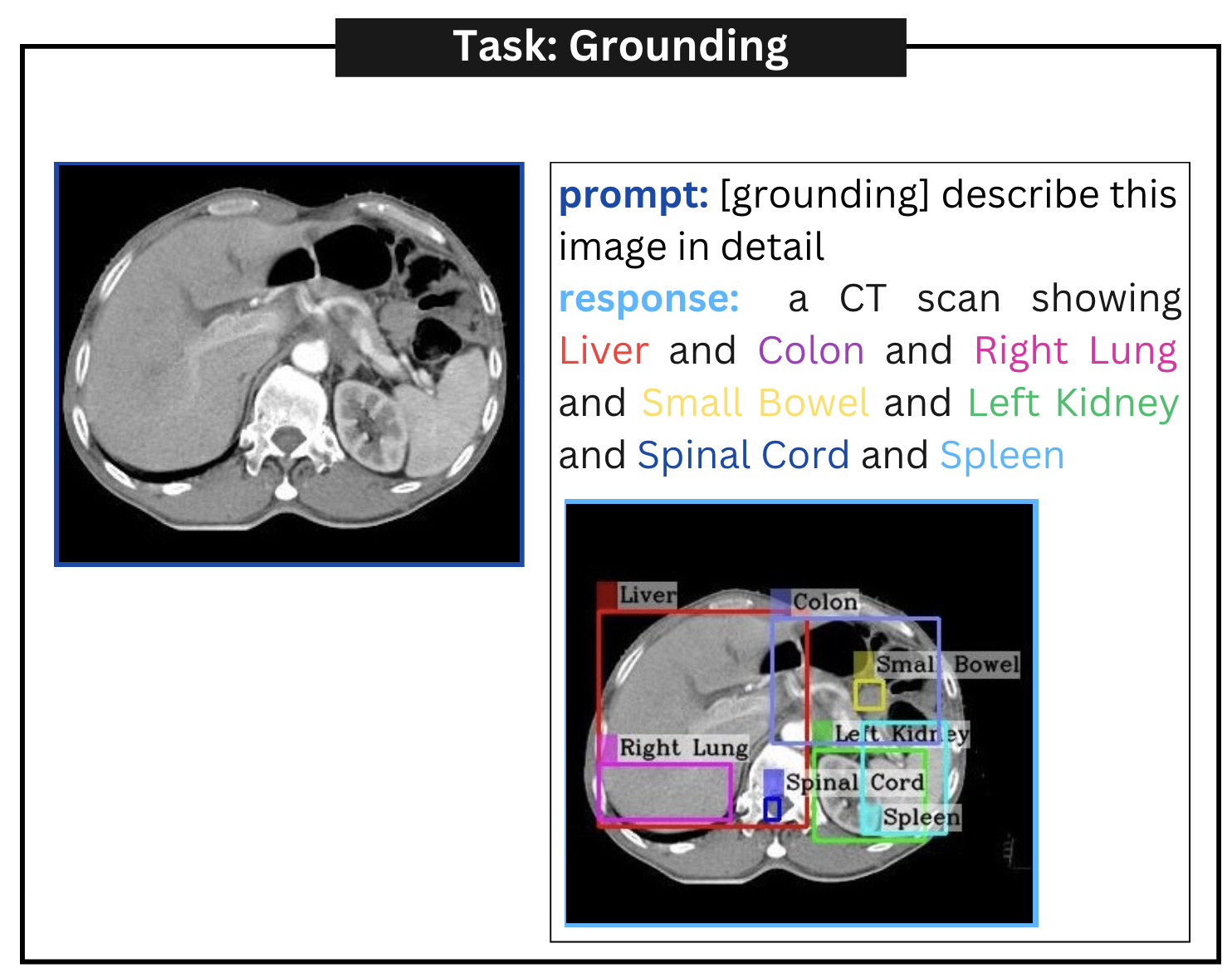}
        \caption{Grounding Task}
        \label{fig:grounding}
        \end{subfigure}
        \hfill
        \begin{subfigure}[b]{0.43\textwidth}
        \centering
        \includegraphics[width=\textwidth]{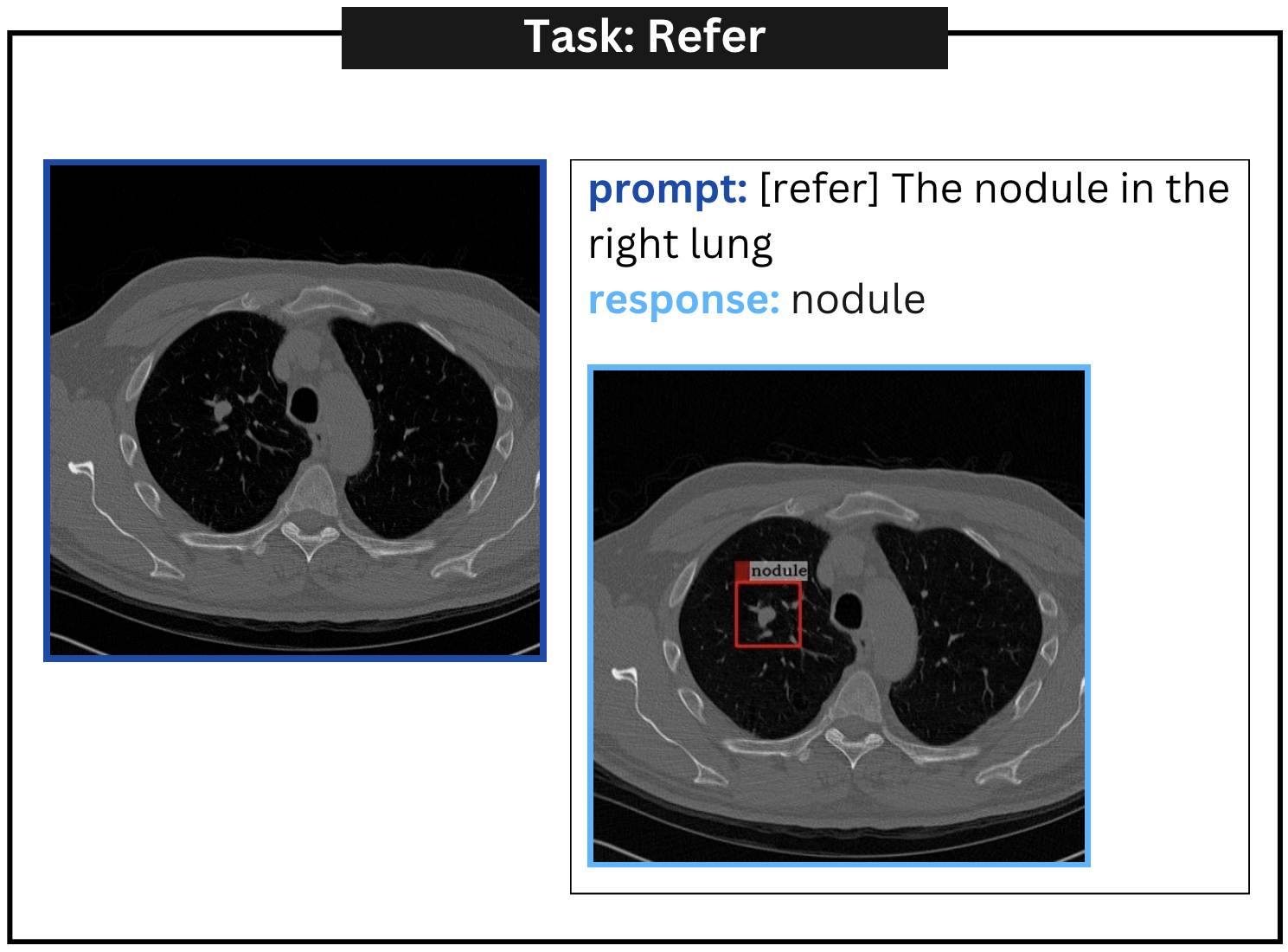}
        \caption{Refer Task}
        \label{fig:refer}
    \end{subfigure}
    \hfill
    \begin{subfigure}[b]{0.43\textwidth}
        \centering
        \includegraphics[width=\textwidth]{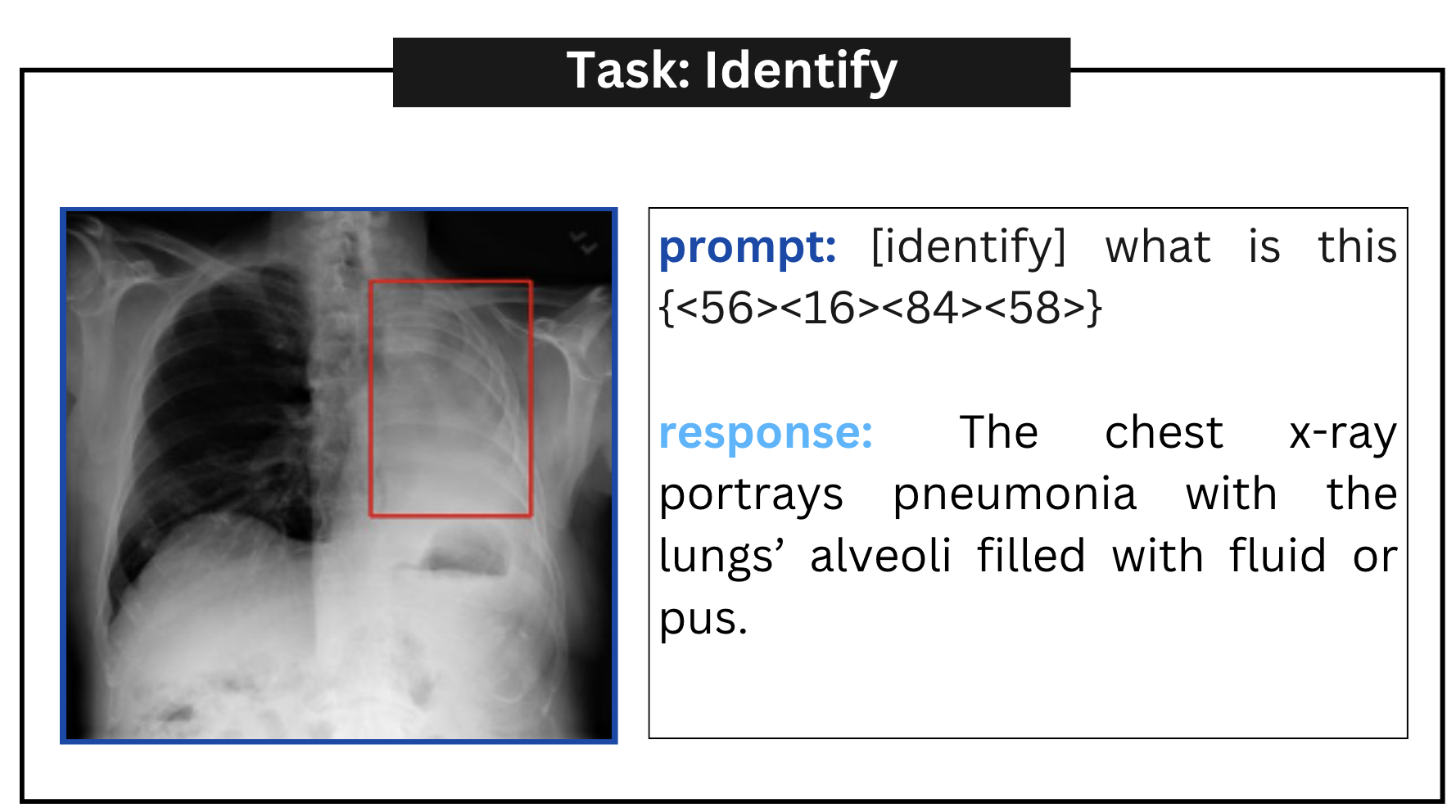}
        \caption{Identify Task}
        \label{fig:identify}
    \end{subfigure}
    \hfill
    \begin{subfigure}[b]{0.43\textwidth}
        \centering
        \includegraphics[width=\textwidth]{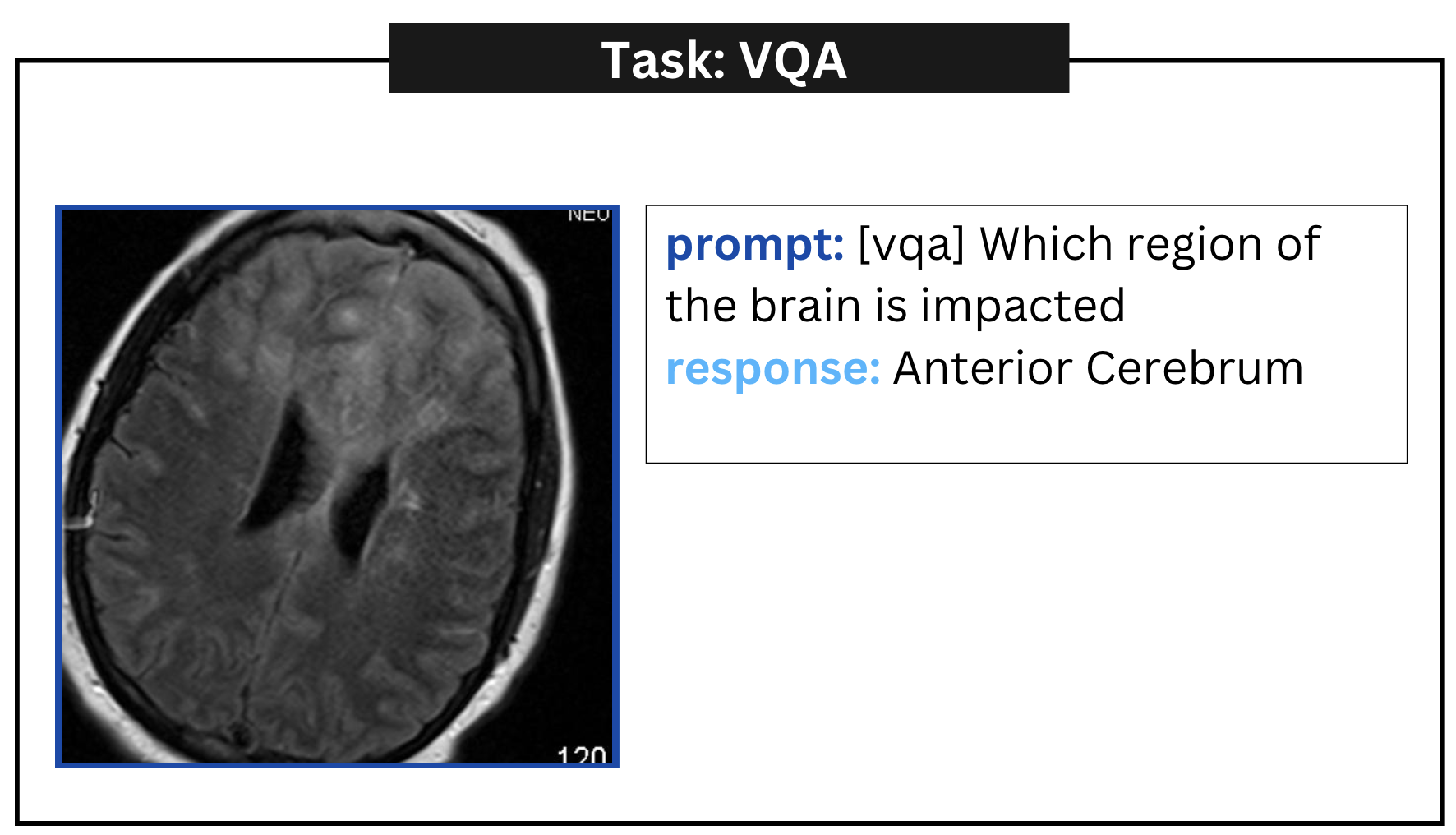}
        \caption{VQA Task}
        \label{fig:vqa}
    \end{subfigure}
    \caption{(Continued) Examples of MiniGPT-Med multi-task abilities include: (c) grounded medical image description, (d) referring disease grounding,(e) identifying diseases, and (f) visual question answering (VQA).}
\end{figure}

\section{False Positive Example}

\begin{figure}[H]
    \centering
    \includegraphics[width=0.43\textwidth]{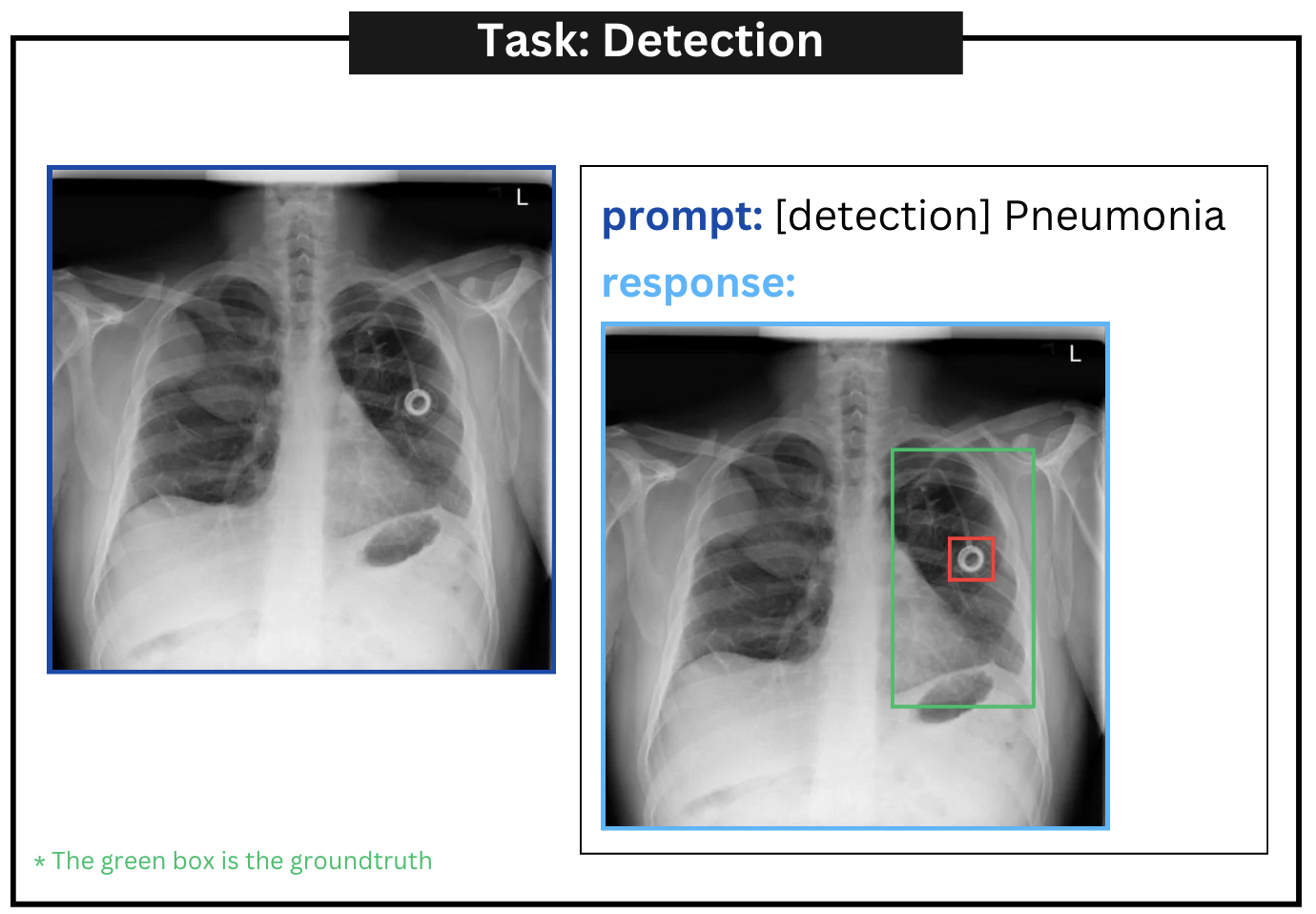}
    \caption{False positive example. The object under the red bounding box denotes the falsely detected disease and the bounding box under the green color represents the ground truth. }
    \label{fig:limitation}
\end{figure}

\end{document}